\documentclass[letterpaper]{article}
\usepackage{aaai}
\usepackage{times}
\usepackage{helvet}
\usepackage{courier}
\usepackage{graphicx}
\usepackage{times}
\usepackage{latexsym}
\usepackage{booktabs}
\usepackage{amsmath}
\usepackage{amssymb}
\usepackage{graphicx}
\usepackage{bm}
\usepackage{CJKutf8}
\usepackage{amsmath}
\usepackage{url}
\usepackage{tabularx}
\usepackage{subfig}
\usepackage{multirow}

\usepackage{color}
\usepackage[ruled,linesnumbered]{algorithm2e}
\begin{document}
%
\title{Financial data analysis application via multi-strategy text processing}
\author{
Hongyin Zhu\\
Department of Computer Science and Technology, Tsinghua University, Beijing, China\\
zhuhongyin2020@mail.tsinghua.edu.cn\\
}
\maketitle
\begin{CJK*}{UTF8}{gbsn}
\begin{abstract}

Maintaining financial system stability is critical to economic development, and early identification of risks and opportunities is essential. The financial industry contains a wide variety of data, such as financial statements, customer information, stock trading data, news, etc. Massive heterogeneous data calls for intelligent algorithms for machines to process and understand. This paper mainly focuses on the stock trading data and news about China A-share companies. We present a financial data analysis application, Financial Quotient Porter, designed to combine textual and numerical data by using a multi-strategy data mining approach. Additionally, we present our efforts and plans in deep learning financial text processing application scenarios using natural language processing (NLP) and knowledge graph (KG) technologies. Based on KG technology, risks and opportunities can be identified from heterogeneous data. NLP technology can be used to extract entities, relations, and events from unstructured text, and analyze market sentiment. Experimental results show market sentiments towards a company and an industry, as well as news-level associations between companies. 
\end{abstract}

\section{Introduction}

Maintaining financial system stability is critical to economic development. To meet this challenge, perceptual and cognitive technologies are necessary to identify risks and opportunities in advance \cite{xiao2021}. Artificial intelligence (AI) can help humans analyze large amounts of financial data in real-time and output conclusions and decisions. Intelligent algorithms are a fundamental part of financial technology (fintech) and are widely used in many application scenarios, e.g., auditing \cite{fisher2016natural}, intelligent early warning \cite{gao2021review,oro2020cognitive}, public opinion monitoring \cite{liu2019new,liu2020research}, quantitative investment \cite{sorensen2019golden}, etc. The financial industry contains a wide variety of data such as company financial statements, customer information, news, stock trading data, industry research reports, etc. In terms of document structure, there are structured, semi-structured, and unstructured data that require different technical solutions.

This paper mainly focuses on applying NLP and KG technologies to textual and numerical data for comprehensive analysis in the financial industry. This paper mainly focuses on the processing of stock trading data and news about China A-share companies. This paper is divided into two parts: the first part introduces an application based on multi-strategy financial data mining, and the second part presents the efforts and plans of our application scenarios of deep learning financial text processing in detail. 

Existing financial NLP has a wide range of applications, such as customer service chatbot \cite{okuda2018ai,quah2019chatbot}, auditing, financial sentiment analysis \cite{araci2019finbert,wang2021qingxu}, public opinion monitoring, intelligent early warning, and stock behavior prediction \cite{DBLP:journals/corr/abs-1912-07700,ris2021index,khedr2017predicting}. However, existing approaches still face many challenges, such as leveraging knowledge graphs to improve language understanding, and using NLP technology to enrich the content of knowledge graphs. Furthermore, the relationship between textual data processing and numerical data mining needs to be further explored. For example, stock trading data, e.g., stock prices, trading volume, Stochastic oscillator (KDJ), moving average convergence/divergence (MACD) and other indicators are purely numerical, while financial news, industry reports, company announcement are textual data. While the gap between numerical and textual data is obvious, they influence each other. On the one hand, news has a clear impact on the stability of the financial system, and on the other hand, stock trading data heralds upcoming news content. Sometimes breaking news events, such as wars, natural disasters, etc., may invalidate data mining techniques for stock trading data. Combining these two types of information, multi-strategy data mining is able to simultaneously consider stock trading data and news from around the world to identify opportunities and risks for conclusions and decisions. 

Combining trading data and financial news to give a comprehensive analysis is a valuable task. The first challenge is the construction of the China A-share companies knowledge graph (CAKG). To resolve this problem, we employ a knowledge graph construction pipeline \cite{zhu2022metaaid}. The second challenge is the heterogeneous data mining of financial data. We propose a multi-strategy data mining pipeline to consider both the trading-level and news-level data. 

Application scenarios of NLP and KG in fintech are extensive. For deep learning financial text processing, KG, NLP, and computer vision (CV) technologies make the machine better understand multi-modal data. KG technology allows the machine to reorganize, link, and understand multi-source heterogeneous data. For financial text processing, entity linking and typing can be used to identify an infinite number of concepts (labels) of entities. Information provenance \cite{hartig2009provenance,lu2018study} and complex network theory \cite{shu2017fake,tang2009qualitative} can be used to verify the reliability or correctness of news. KG reasoning can be used for conflict verification and to discover new opportunities and risks. Users can obtain answers to complex financial questions through KG question answering (QA). 

NLP technology allows machines directly understand the unstructured text or extract structured key information. For financial text processing, NLP can be used to classify company announcements, news, and user comments into different sentiment categories (positive, neutral, negative, etc.) and different levels of credibility. Extracting entities, relations, and company events of interest from the plain text will enable machines to analyze key information. Machine reading comprehension returns satisfactory answers from massive documents. Text clustering can be used to compare the data across industries. Abstractive summarization helps investors and analysts read documents more efficiently. The main contributions of this paper are as follows.


1. We propose a multi-strategy financial data mining pipeline to analyze numerical and textual data. 

2. We develop an application to process stock trading data and news and evaluate company performance from multiple dimensions. 

3. We present our efforts and plans in deep learning financial text processing application scenarios, including challenges and methods for the wider use of NLP and KG in financial data analysis. 

\section{Related Work}
For financial NLP, Araci \cite{araci2019finbert} pre-train a FinBERT on a large financial corpus and a small task-specific corpus for sentiment analysis which help to classify how the markets will react to the information presented in the text. Liu et al. \cite{DBLP:conf/ijcai/0001HH0Z20} pre-train a FinBERT to improve the performance on financial text understanding. This model is trained on 6 self-supervised tasks on financial corpora. Yang et al. \cite{yang2020finbert} pre-train a FinBERT on a large financial communication corpus and achieve performance improvements on sentiment analysis. Khan et al. \cite{khan2021artificial} propose a chatbot architecture to deal with the business in the finance and banking industry. Yıldırım et al. \cite{yildirim2018classification} propose a machine learning pipeline to resolve the problem of classifying the financial news as significant and nonsignificant. 

For financial KG, Alam et al. \cite{alam2022loan} propose to combine KG and machine learning to improve the prediction performance of loan default risk. They build a KG to train the KG embeddings to use as input to the model. Huai et al. \cite{elhammadi2020high} propose a high-precision knowledge extraction pipeline for extracting key information in financial news, in which they use semantic role labels and a conditional random field (CRF) model to identify semantic relationships between entities in noisy text. Cheng et al. \cite{cheng2020knowledge} propose to use KG-based event embeddings for quantitative investment by jointly training FinKGs, events, and relations and feeding them into models to derive investment strategies. Zehra et al. \cite{zehra2021financial} build a financial report query system to query annual financial reports to help investors in the banking sector. They extract information from annual reports to build KGs and use ontologies to help understand user queries. Liao et al. \cite{liaokuo2020} construct a financial event graph by extracting causal relationships between events and use this technique for stock market forecasting. 

For financial data mining, Zhang et al. \cite{zhang2020autoalpha} propose a hierarchical evolutionary algorithm to locate promising search spaces and mine alpha factors in quantitative investment \cite{sorensen2021quantitative}. Cheng et al. \cite{cheng2021establishing} present the way they combine decision tree and Apriori algorithm \cite{agrawal1996fast,han2000mining} into investment decision models. Chang et al. \cite{chang2021stock} present a data mining pipeline based on neural network, support vector machine, mixed data sampling, etc. for stock price prediction. Kim \cite{kim2021data} proposes a financial market data mining framework, including the whole process of data preprocessing, feature selection, model, evaluation, and reporting. Li et al. \cite{li2021integrated} improve clustering algorithms by proposing a criterion based on spectral graph theory to evaluate the cluster quality. Researchers \cite{al2021financial,sanad2021financial} review the data mining technology for financial fraud detection. 

\section{Approach}
In this section, we first introduce our application in multi-strategy financial data mining. Then we present the application scenarios of KG and NLP techniques in deep learning financial text mining.

\subsection{Multi-strategy Financial Data Mining}
Multi-strategy financial data mining aims to use multi-strategy data analysis algorithms to comprehensively analyze numerical data and textual data to obtain market sentiments of companies and industries, such as food and beverage, semiconductor industry, lithium battery industry, etc. Finally, we can infer whether company performance or target industry performance is in line with expectations.

We present a mobile app ``financial quotient porter'' (财商数据) which aims to analyze the stock trading data from multiple perspectives. Share prices typically reflect a company's current quarter performance or future growth potential. To estimate company performance, we first construct a China A-share companies knowledge graph to support the analysis of stock trading data. We propose a multi-dimensional scoring (MDS) pipeline which is designed to analyze a company from multiple dimensions. We assign different weights to different dimensions, for example, market sentiment towards a company, sentiment comparison among different investment groups, joint analysis of sentiment among multiple investment groups, etc. We further analyze the sentiment of different industries and estimate whether the development of these industries is in line with expectations.

News is a non-negligible factor in maintaining financial stability because the stock market is sensitive to news. Many stocks go up by the limit when there is good news, and down by the limit when there is bad news. Some stocks react to the news too quickly for the market to identify opportunities or avoid risks. If investors want to discover indirect opportunities or avoid indirect risks, the news-level association can help find closely related companies. News-level associations are time-sensitive, meaning two companies weren't connected a week ago, but now they're connected through an event, and their connection may disappear a month later. We can discover news-level related companies and conduct a comprehensive analysis of their trading data. To find news-level related companies, we employ information extraction techniques \cite{zhu2022metaaid}. We construct a weighted graph based on the extracted structured information to analyze the degree of news-level association between companies. Different from static relations such as (company A, holding, company B), (person X, directorOf, company A) \& (person X, directorOf, company B), we focus on the real-time news-level associations rather than relations that have always existed in the past. 

\subsection{Deep Learning Financial Text Processing}
We present our application scenarios of deep learning financial text processing with our efforts and plans to identify financial opportunities and risks using NLP and KG techniques. Note that this subsection mainly describes the application scenarios, but deep learning financial text processing is not limited to this. Many technologies not mentioned are worth exploring further application scenarios.

\subsubsection{Financial KG Applications}
News or intelligence has a great impact on the stability of financial markets, and early identification or intelligence inference is crucial for finding opportunities and avoiding risks. The presence of undetected entities in news may cause problems, requiring us to identify complete entities as much as possible. For example, given the sentence ``Due to covid-19, 1664 will reduce sales by x\% in 2022.", we humans can simply identify the entity ``1664", i.e., the beer brand, craft beer, and many people may not know which company this brand belongs to. Entity linking (EL) \cite{sevgili2020neural,shen2014entity}, also known as named entity disambiguation (NED), includes both named entity recognition and disambiguation processes. The challenges lie in irregular entity mentions, long-tail entities, and entity ambiguity. Although named entity recognition (NER) models achieve state-of-the-art performance on some datasets, they can only recognize limited entities and require large amounts of high-quality training data. If the word ``1664" was filtered out as a meaningless number, the system would not find news about this brand. With a knowledge graph containing rich entities, we can identify complete investment-related entities from plain text using an entity linking pipeline.

Entity disambiguation is important for accurately targeting the market segments (subdivision), which is the second challenge of entity linking. Many entities have the name ``1664'', such as \texttt{E1:/year (/年份), E2:/food\_and\_beverages/beer (/食品饮料/啤酒), E3:/brand/beer\_brand (/品牌/啤酒品牌)}, etc. Using a context-aware entity disambiguation algorithm, we can link ``1664'' to the correct entity \texttt{E3:/brand/beer\_brand}. Obscure entities are more challenging. For example, ``The data shows that the net profit of Great Wall fell by x\% year-on-year, which may be related to the upgrade of the energy structure" where ``Great Wall" maybe \texttt{E4:/auto (/汽车), E5:/auto/new\_energy (/汽车/新能源), E6:/auto/traditional\_energy (/汽车/传统能源), E7:/company/China\_Great\_Wall (/公司/中国长城), E8:/company/ Great\_Wall\_ Motors (/公司/长城汽车), E9:/travel/tourist\_attraction/the\_Great\_Wall (/旅游/景点/长城)}, etc. With the help of KG, this model can infer the correct entity by reading the ``energy structure" in the context, and conclude this ``Great Wall'' should be the \texttt{E6:/auto/traditional\_energy, /company/Great\_Wall\_Motors}. 

Assigning sub-industry concepts to entities facilitates reasoning for better decision-making. Let's still take the above example, ``Due to covid-19, 1664 will reduce sales by x\% in 2022.", we humans know that the fine-grained labels for ``1664'' are \texttt{/company/Chongqing\_Beer, /alcoholic\_beverages/beer, /alcoholic\_beverages/beer/craft\_beer, /brand/beer\_brand, /food\_and\_beverages}, etc. From these labels, we can infer that this is risky news for the \texttt{/company/Chongqing\_Beer (重庆啤酒)}. At the same time, this is risky news for other products belonging to \texttt{/alcohol\_beverages/beer/craft\_beer}, such as ``Corona Beer". Nevertheless, that does not mean it is risky news for \texttt{/alcoholic\_beverages/liquor} or \texttt{/company/Kweichow\_Moutai}. Fine-grained entity-typing \cite{murty2018hierarchical,onoe2020fine,DBLP:conf/emnlp/LiuLXH0W21} aims to assign fine-grained concepts to entities. The challenge lies in the large number of concepts and the concept hierarchy. Prompt-learning \cite{DBLP:journals/corr/abs-2108-10604} achieves state-of-the-art performance on some datasets. For financial application scenarios, there will be more concepts beyond the capacity of a single model, and the use of knowledge graphs and ontologies can help address this problem, allowing machines to more accurately infer information about opportunities and risks. 


News spreads like Wildfire, but its reliability and correctness are difficult to evaluate. For example, ``Pork prices will increase by x\% in 2 months". Whether this outgoing news is reliable is difficult to assess, but investment opportunities often come with risks. Knowledge validation from heterogeneous data is an important technology to verify the reliability of the information. The challenge lies in the way to find evidence and track news sources (provenance \cite{hartig2009provenance}). It is a good choice to verify the information from the perspective of graph structure (complex network) \cite{zhou2019network,alassad2019finding}. Tracking the platforms where news is published and the path of news dissemination on the Internet, then the reliability of news can be analyzed according to the credibility of each node and the graph structure. 

When dealing with large amounts of company performance data, analysts need to ensure data consistency. For example, extracting from different files we got the performance of Q1 result $>$ Q2 result $>$ Q3 result $>$ Q1 result (第一季度$>$第二季度$>$第三季度$>$第一季度) which is contradictory. The out-of-range attribute value is also an error, and it is difficult to manually check for implicit errors. Knowledge validation aims to improve knowledge quality \cite{owoc1999principles}. Identifying conflicts in data is an important aspect of KG reasoning, and there are some knowledge reasoning tools \cite{vermesan2013validation}. Whether it is an obvious data error or an implicit property range error, the location of the error can be tracked down.

Investors need to infer new opportunities and risks based on existing data. For example, (Company A, business partner, Company B), (Company B, suffer, huge fine) $\rightarrow$ (Company A, stock price, falling).  (Company A, holding, Company B), (Company B, performance, +120\%) $\rightarrow$ (Company A, stock price, rising). Knowledge reasoning \cite{chen2020review,DBLP:journals/corr/abs-2108-06040} aims to identify errors and infer new facts from existing data. The challenge lies in inference rule mining and interpretability of inference results. Inference algorithms can be divided into rule mining \cite{galarraga2013amie,lao2011random}, reinforcement learning \cite{DBLP:conf/emnlp/XiongHW17}, knowledge representation learning \cite{saxena2020improving,DBLP:conf/nips/BordesUGWY13}, etc. By mining rules or having experts design them, machines can infer opportunities and risks through knowledge graphs.

Querying company data based on complex questions is a common task. For example, ``List the top 5 companies with the fastest fourth quarter \textbf{growth} of all companies producing premium beers" (列出所有生产高端啤酒的公司中第四个季度业绩\textbf{增速}最高的前5名的公司). While investors can obtain this data from search engines or stock brokers websites (such as 10jqka.com.cn, eastmoney.com, etc.), managing this data is time-consuming. Knowledge graph question answering (KGQA) \cite{huang2019knowledge,hao2017end} aims to find answers to natural language questions over a knowledge graph. Investors can ask complex questions to the Knowledge Graph. The challenges lie in the completeness of the knowledge graph, the semantic parsing of questions, and the reasoning of answers. While the query language for interacting with knowledge graphs is SPARQL, the system should have the ability to parse natural language questions into SPARQL \cite{yih2016value,cao2020kqa}. Deep learning models have achieved good performance in translating natural language question into SPARQL \cite{yin2021neural,DBLP:journals/corr/abs-1803-04329}. 


\subsubsection{Financial NLP Applications}
News can be classified into different degrees of good news or bad news according to sentiment, or news can be divided into credible news, medium credible news, untrustworthy news, and junk news according to credibility. It is important to analyze the sentiment of financial news or the sentiment of user comments. Text classification \cite{kowsari2019text,DBLP:conf/emnlp/Kim14,yang2016hierarchical} aims to assigns a set of predefined categories to the documents. The challenge lies in that the same content can be expressed in different ways, and different types of documents express sentiment in different ways. In terms of sentiment classification \cite{zhang2018deep,yadav2020sentiment}, documents of different genres have different target categories and their expressions are different. For example, news expresses sentiment objectively, user comments express sentiment directly, and financial reports express sentiment implicitly. Different models can be trained on different types of documents to predict the sentiment distribution of events, companies, industries, etc.


Structured text only accounts for a small proportion, and the content in massive unstructured text is underutilized. Extracting structured knowledge from the unstructured text can be used to complete knowledge graphs for better services. Information extraction (IE) \cite{deng2018deep,miwa-bansal-2016-end} includes two subtasks, named entity recognition (NER) \cite{DBLP:conf/acl/MaH16,DBLP:conf/naacl/LampleBSKD16} and relation extraction (RE) \cite{DBLP:conf/emnlp/HanGYYLS19,DBLP:conf/acl/YaoYLHLLLHZS19}. NER is used to identify entities in unstructured text, and RE is used to extract relations between entities. NER faces many challenges, such as nested entities, high-frequency irregular entity abbreviations, long-tail entities, etc. RE is divided into supervised RE, distant-supervision RE and open IE \cite{mausam2016open}. Supervised RE involves the definition of pre-defined relations, data annotation, model training, and evaluation. Distant-supervision (DS) RE uses relations in KGs to automatically annotate corpora. This method can conduct relation extraction and performance evaluation without manual annotation of the corpus. However, DSRE faces the problem of noisy samples in the training data. Open IE does not require pre-defined relations and extracts entity relations based on various strategies such as statistics, dictionaries, rules, syntactic parsing, word vectors, etc. The challenges lie in quality control, entity relation normalization, etc. Graph search algorithms can be used on the extracted graphs to identify market opportunities and risks.

Various events occur every day around the world, some of which affect the current and future prices of the stocks of related companies. For example, ``Due to covid-19, 1664 will decrease sales by x\% in 2022", the event is ``decrease sales", and the event arguments are (company: Chongqing beer), (amount: decrease x\%) and (cause: covid-19). Extracting events \cite{DBLP:conf/emnlp/WangWHJHLLLLZ20} from documents can build an event graph to analyze the opportunities and risks. Event extraction \cite{xiang2019survey} aims to extract structured event types and arguments from unstructured text. The challenge lies in that we encounter many types of events in the financial market and the event arguments have various expressions contexts. Furthermore, manual labeling of training data requires summarizing and annotating events in unstructured text, which is labor-intensive. Pipeline or joint extraction \cite{sha2018jointly} models can be used to extract event triggers and arguments. 

``After the price increase of Kweichow Moutai, what happened to the prices of other brands of liquor?'' (贵州茅台涨价的时候其他品牌的白酒的价格怎么变化的？) When we want to know something about financial markets, we usually use a search engine, which returns web pages from open domains and requires manual sifting of the web pages. Information retrieval (IR) based question answering \cite{abbasiantaeb2021text,zhu2021collaborative} aims at finding short passages for users' questions. The answers are directly searched from unstructured documents. This task consists of two stages, coarse ranking, and reranking. The challenge lies in the semantic matching of question answering pairs. Pre-trained language models achieve state-of-the-art performance on this task. Question answering systems allow users to get concrete answers directly.

With the booming of news, reports, social media, self-media, etc., massive documents need to be classified and organized based on their key information such as policies and companies in each cluster. Dividing documents into different groups for different market segments, policies, and industries can be used for fine-grained comparison and analysis. Text clustering aims to group a set of unlabeled texts with high similarity in the same group and low similarity in different groups. Different clusters can also be compared to each other. The challenge lies in the representation of the document and the clustering algorithms. Pre-trained language model achieves state-of-the-art performance to encode documents. Clustering algorithms \cite{bishop:2006:PRML} use hierarchical clustering, k-means, DBSCAN \cite{ester1996density,schubert2017dbscan}, etc. to group documents into hierarchical or single-level clusters. 

Massive unstructured reports are not adequately read. Investors need machines to compress text length to improve reading efficiency. Text summarization aims to condense long documents into a shorter version while preserving the key information and meaning of the content. The challenge lies in the summary generation and evaluation. The deep learning model \cite{el2021automatic} has achieved good performance in generating concise and accurate abstracts.

Many natural language technologies are not used as separate application scenarios, but are used in different stages of KG and NLP applications, e.g., word vectors \cite{vrehuuvrek2011gensim}, pre-trained language models \cite{DBLP:conf/naacl/DevlinCLT19}, word segmentation, part-of-speech tagging, syntax parsing \cite{manning2014stanford,bird2004nltk}, etc.

\section{Experiment}
In this section, we start by analyzing the sentiments of different groups towards the company. We then analyze the sentiments of different groups towards an industry. Finally, we show news-level correlations between companies. 

\subsection{Setup}
We run this app on an iPhone XR. The cloud services run on an AMD Ryzen 5 1500X Quad-Core Processor @ 3.5GHz (Mem: 16G) and 1 Tesla T4 GPU (16G). 
\subsection{Results of Company Analysis}
\begin{figure}[htbp!]
\centering
\includegraphics[width=3.2in]{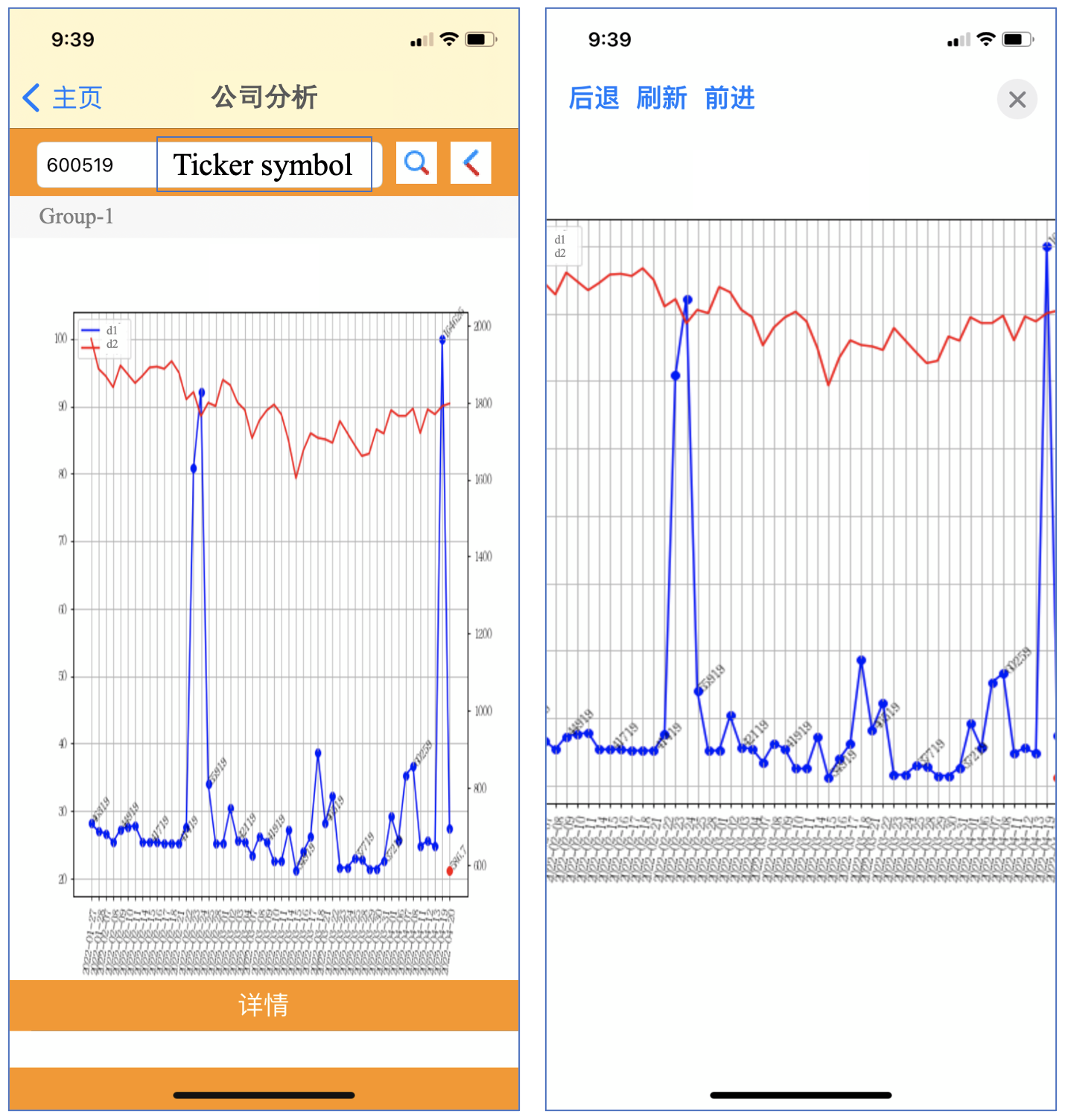}
\caption{Sentiment variation of Group-1 towards Kweichow Moutai\label{gegu1}}
\end{figure}

\begin{figure}[htbp!]
\centering
\includegraphics[width=3.2in]{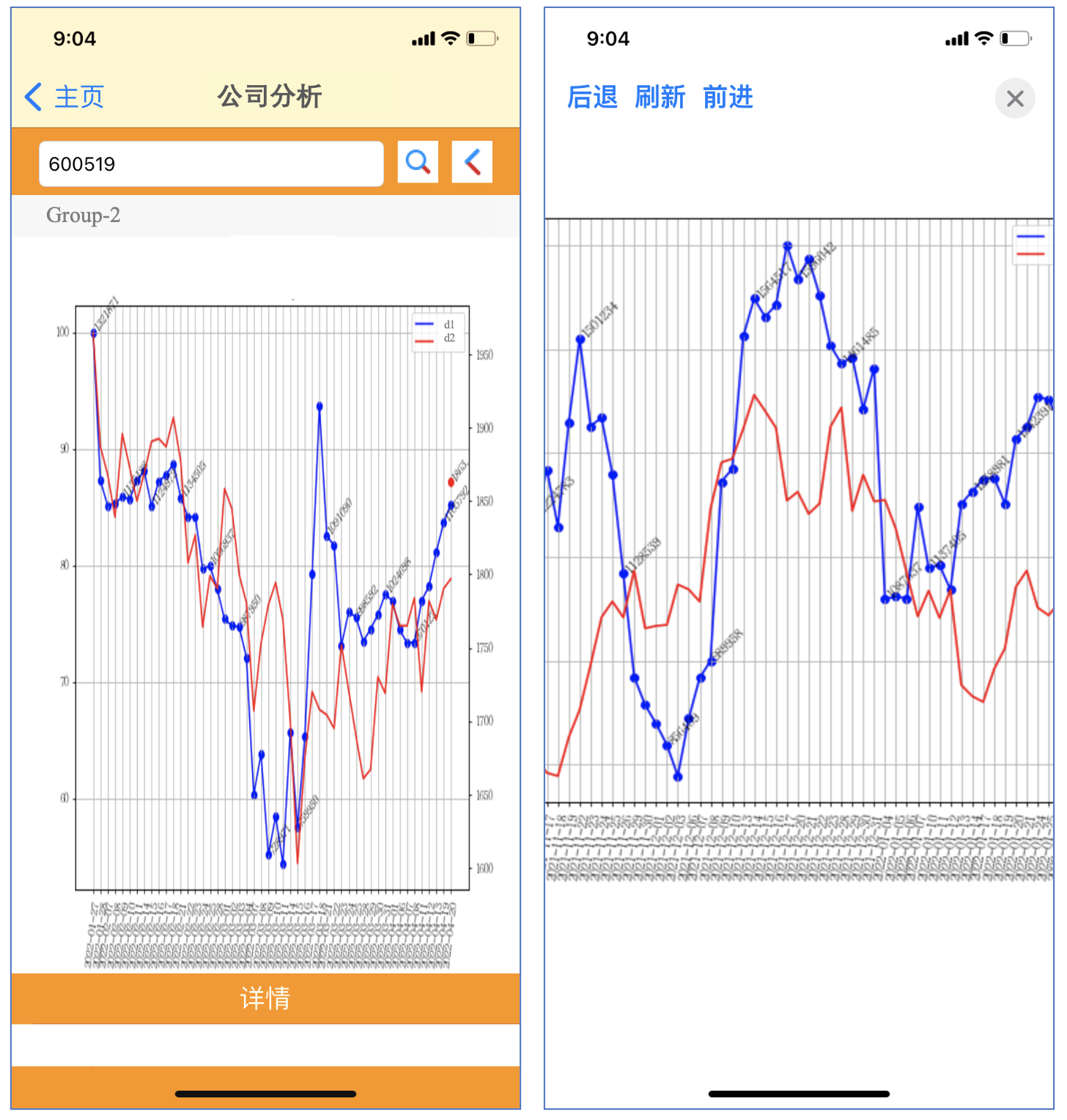}
\caption{Sentiment variation of Group-2 towards Kweichow Moutai\label{gegu2}}
\end{figure}

Figure \ref{gegu1} and Figure \ref{gegu2} show the market sentiment analysis of Kweichow Moutai (贵州茅台) from February 2022 to April 2022, where the blue and red lines denote the sentiment of different groups in the market and stock prices respectively. Figure \ref{gegu1} shows the sentiment of group-1 towards Kweichow Moutai. Their views on Kweichow Moutai are relatively stable, with only noticeable changes in sentiment on specific dates such as February 23, 2022, and April 19, 2022. Figure \ref{gegu2} shows the sentiment of Group-2 towards Kweichow Moutai. 
We can see that their sentiment are more sensitive. For example, on March 11, 2022, their sentiment reached a minimum, and on March 18, 2022, their sentiment reached a maximum. Overall, the sentiment of Kweichow Moutai is still divided, but not too bad. 

\subsection{Results of Industry Analysis}
\begin{figure}[htbp!]
\centering
\includegraphics[width=3.2in]{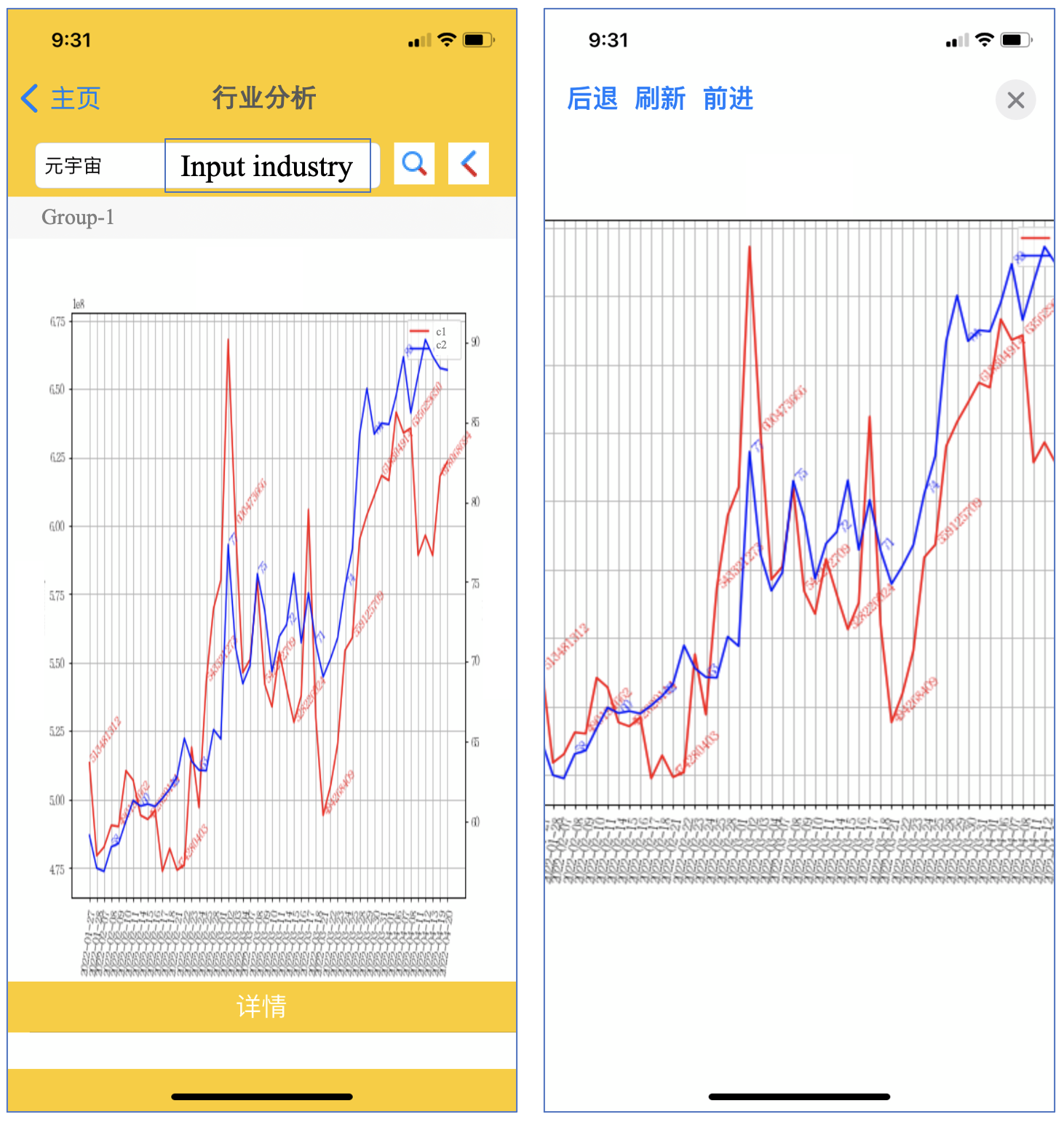}
\caption{Sentiment variation of Group-1 towards metaverse industry\label{hangye1}}
\end{figure}

\begin{figure}[htbp!]
\centering
\includegraphics[width=3.2in]{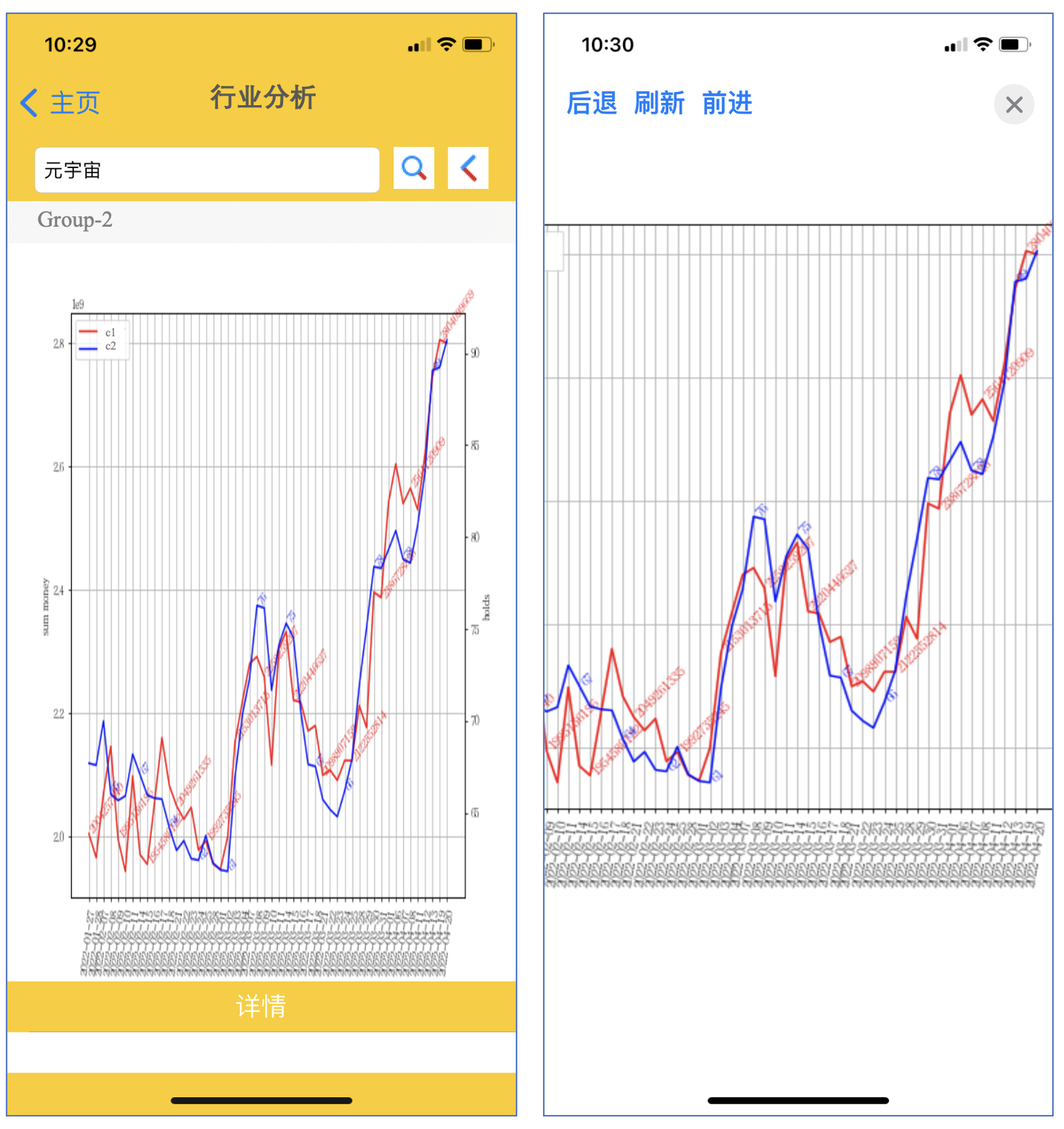}
\caption{Sentiment variation of Group-2 towards metaverse industry\label{hangye2}}
\end{figure}

Figure \ref{hangye1} and Figure \ref{hangye2} show the market sentiment analysis for the metaverse industry from February 2022 to April 2022, where the blue line denotes the sentiment of different groups in the market and the red line represents a comprehensive indicator, taking into account market sentiment and the actual performance of the industry. As shown in Figure \ref{hangye1}, from March 18, 2022, to April 20, 2022, positive sentiment continued to rise. However, the red line is very unstable. For example, a maximum value was reached on March 2, the market sentiment retreated immediately, and a minimum value was reached on March 18. This means the actual performance of the metaverse industry is not ideal. As shown in Figure \ref{hangye2}, the sentiment of Group-2 towards the metaverse continued to grow from March 23rd. Unlike Group-1, the red line of Group-2 continued to rise. This is because the positive sentiment level of Group-2 is higher than that of Group-1, so the upward trend of the red line is relatively stable. 

\subsection{Results of News-level Associations}
\begin{figure}[htbp!]
\centering
\includegraphics[width=3.2in]{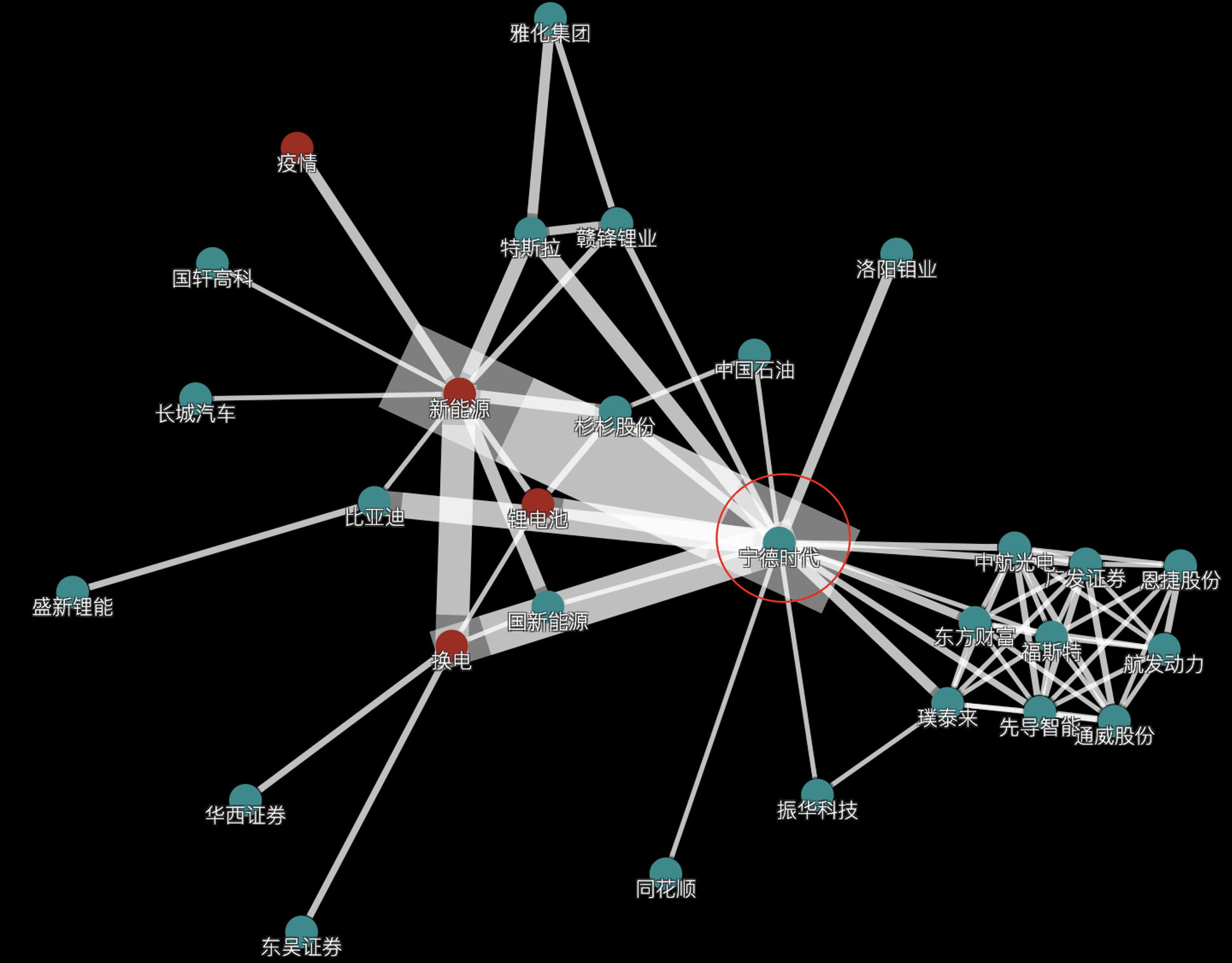}
\caption{Companies with news-level associations to CATL\label{newslevel}}
\end{figure}

We take Contemporary Amperex Technology (宁德时代, CATL) as an example to analyze the relationships between companies on April 21, 2022. Figure \ref{newslevel} shows a graph of news-level associations between companies centered on CATL, where CATL is marked with a red circle. Red nodes represent industry concepts and green nodes represent companies. Edges represent relationships between nodes, and edge thicknesses represent connection weights. We can see that concepts directly related to CATL include new energy (新能源), lithium batteries (锂电池), and power exchange (换电). The concept pandemic (疫情) is indirectly related to the CATL through new energy, indicating that the pandemic has an impact on the CATL through the new energy industry. Companies directly related to CATL, in descending order of weight, are BYD (比亚迪), TSLA (特斯拉), Ningbo Shanshan (杉杉股份), China Molybdenum (洛阳钼业), Shanghai Putailai New Energy Technology (璞泰来), GanFeng Lithium (赣锋锂业), Yunnan Energy New Material (恩捷股份), Hangzhou First PV Material (福斯特), China Zhenhua (Group) Science \& Technology (振华科技), AVIC Jonhon Optronic Technology (中航光电), etc.

\section{Conclusion and Future Work}
This paper introduces a financial data analysis app that combines stock trading data and news for comprehensive analysis and decision making. We propose a multi-strategy financial data mining pipeline and news-level associations. Besides, we present our deep learning financial text processing application scenarios. We describe our efforts and plans to uncover financial opportunities and risks using NLP and KG technologies. In the future, we hope to resolve the problem of unified representation of numerical and textual data and apply it to quantitative investment, while expanding the application scenarios of NLP and KG in financial data.

\end{CJK*}
\bibliographystyle{aaai}
\bibliography{reference}

\end{document}